\documentclass[runningheads]{llncs}
\usepackage{graphicx}
\usepackage{cite}
\usepackage{hyperref}
\usepackage{times}
\usepackage{latexsym}
\begin{document}
\title{Ensuring the Inclusive Use of Natural Language Processing in the Global Response to COVID-19}
\titlerunning{Ensuring the Inclusive Use of NLP in the Global Response to COVID-19}
%
\author{Alexandra Sasha Luccioni\inst{1,2} \and
Katherine Hoffmann Pham\inst{1,3} \and
Cynthia Sin Nga Lam\inst{1}\and Joseph Aylett-Bullock\inst{1,4} \and Miguel Luengo-Oroz\inst{1}}
\authorrunning{A.Luccioni et al.}
%
\institute{United Nations Global Pulse \email{\{sasha,katherine,cynthia,joseph,miguel\}@unglobalpulse.org} \and Mila Qu\'{e}bec Artificial Intelligence Institute, Universit\'{e} de Montr\'{e}al, Canada \and NYU Stern School of Business, New York, USA \and Institute for Data Science, Durham University, United Kingdom}
 
%
%
\maketitle              
\begin{abstract}
Natural language processing (NLP) plays a significant role in tools for the COVID-19 pandemic response, from detecting misinformation on social media to  helping to provide accurate clinical information or summarizing scientific research. However, the approaches developed thus far have not benefited all populations, regions or languages equally. We discuss ways in which current and future NLP approaches can be made more inclusive by covering low-resource languages, including alternative modalities, leveraging out-of-the-box tools and forming meaningful partnerships. We suggest several future directions for researchers interested in maximizing the positive societal impacts of NLP. 
\end{abstract}
\section{Introduction and Context}
The COVID-19 pandemic has changed the way we live, work, and travel. In their response to the pandemic, governments and organizations have grappled with pandemic response, deploying numerous tools and initiatives, many of which leverage novel technologies to guide decision-making and build more resilient communities. Textual resources in particular have been critical in pandemic response, facilitating access to scientific information to guide research and enabling the detection of misinformation in the midst of the COVID-19 `infodemic'~\cite{bullock2020mapping, Eysenbach2020how}. Going forward, there is a continuing need for the use of NLP to facilitate the search for and dissemination of accurate information and to counter false narratives and vaccine skepticism.
However, the availability, accessibility and use of these resources has been unequal, prioritizing certain languages and benefiting some communities more than others, exacerbating existing inequalities and widening the gap between those who have access to advanced digital technologies and those who do not~\cite{beaunoyer2020covid}. 

To ensure the inclusive use of NLP in the global response to the COVID-19 pandemic, we argue that it is paramount to develop NLP tools that are both comprehensive and accessible, and to apply them in ways that maximize positive impact and minimize potential risks. In this commentary, we identify four areas that we find to be particularly important to achieving this goal: covering low-resource languages, including alternative modalities such as audio or video, using out-of-the-box tools where appropriate and forming meaningful partnerships to broaden access to technical and local expertise, as well as data and tools. We discuss each of these themes below, and conclude with a series of observations and considerations for future research.

\section{Low-Resource Languages}\label{sec:LRL}

While the exact number of existing languages is debated, estimates range anywhere from 6,000 to 7,000~\cite{languages2020}. However, current NLP technologies such as language models, datasets and tool kits have predominantly been developed for a limited number of `high-resource' languages, which can leave large parts of the world behind. In the context of the COVID-19 pandemic response, where fighting the infodemic of misinformation and fake news has become a top priority, low-resource languages may also suffer a disadvantage since accurate information from verifiable sources can arrive with delays and gaps. Furthermore, while setting up a simple pipeline for detecting misinformation in English or French can take a few days using pre-trained models and tools, it can quickly become a multi-month, resource-intensive endeavor for underrepresented languages~\cite{bullock2019automated}. Finally although research initiatives for misinformation detection in low-resource languages do exist (e.g.~\cite{hossain2020banfakenews, cruz2019localization}), there have not been, to our knowledge, tools deployed for this purpose in the context of the COVID-19 pandemic. 

 There are many challenges to developing NLP tools for low-resource languages, ranging from 
 regional variants that are hard to represent using standard language models (e.g. Arabic), to languages with predominantly oral traditions that have little or no written texts (e.g. Métis). However, the biggest bottleneck for developing language models and tools for low-resource languages remains data availability. For example, mainstream NLP approaches such as word embeddings and large language models are difficult to implement since there are few pre-compiled large text corpora available~\cite{magueresse2020low}. This has spurred promising research on NLP in low-data scenarios, with approaches ranging from few-shot and transfer learning~\cite{johnson2017google, levy2020cross, spangher2020enabling} to data augmentation~\cite{fadaee2017data} and cross-lingual language models~\cite{conneau2020unsupervised}. However, many languages remain under-served in terms of NLP resources and datasets, especially in the context of misinformation detection, which often requires customizing neural language models to encompass specific contexts and topics. In addition to further research developments in this area, it is important to promote the work of those teams compiling large datasets for previously low-resource languages~\cite{orife2020masakhane, scannell2007crubadan}. Overall, the state of available resources for languages that are not mainstream is unfortunate and makes it difficult to deploy scalable NLP systems for these languages. Finally, more transparency should be established between existing commercial endeavors to detect fake news (such as those deployed by major social media websites) and researchers, since releasing datasets that have been flagged as misinformation by human moderators can help train NLP models to mimic this behavior.  

\section{Alternative Modalities} \label{sec:modalities}
In addition to improving the coverage of underrepresented languages, analyzing data from a broader range of modalities could also help to expand the scope of NLP research. Given the need for rapid response to the COVID-19 pandemic, there has been a bias towards analysis of text or multimedia data sources that are easy to mine, such as Twitter. However, these types of content are created and consumed by only a fraction of the world's population, and are biased towards those with ready access to the internet. In other contexts, alternative channels such as radio, television, or messaging applications might be more common sources of information \cite{newman2020key}. 

Radio in particular is a promising channel for studying opinions and discussions of emerging topics such as COVID-19. There are an estimated 44,000 radio stations worldwide (see e.g. \url{http://radio.garden} for a sample), and in the developing world radio stations typically have far wider and more consistent coverage than high-speed internet, with radios available to 75\% of
households~\cite{unesco_statistics_2013}. Even in the United States, radios were estimated to reach 92\% of adult listeners on a weekly basis as of 2019, exceeding the reach of any other channel~\cite{nielsen_steady_2019}.
Radio shows disseminate news, discuss current events, debate rumors, record spontaneous expressions of local opinions and capture information-seeking behavior. In the face of crises or natural disasters, radio may be one of the quickest and most reliable sources of information about the situation on the ground~\cite{munro_crowdsourcing_2013}. 

During the COVID-19 pandemic, researchers at the United Nations (UN) have used radio mining techniques in Uganda to monitor discussions relating to outbreaks and health system capacity, rumors and misinformation, and the socioeconomic impacts of COVID-19~\cite{hidalgo-sanchis_using_2020}. 
Developing a radio-based NLP pipeline poses a number of distinct challenges. 
These include engineering challenges such as capturing and recording radio streams; efficiently managing the high volume of data collected; and transcribing audio recordings to text for further analysis. They also include technical challenges such as speaker disambiguation when multiple individuals are interacting in conversation; filtering out irrelevant content (e.g. commercials, news, and music); and identifying topic-relevant discussions through approximate keyword searches (for example, transcription may introduce new keyword variants -- such as ``Koby'' and ``cobi'' when ``COVID'' is said by Spanish speakers). Given the relative lack of prior research on radio data, these challenges might pose fruitful areas for innovation in the NLP community.

Radio is just one of several possible alternative modalities; as noted above, other data sources of interest include television and informal messaging channels or even SMS-based question-and-answer forums.
In addition to original research analyzing these channels, another potential direction for future work involves the creation of tools to better enable on-the-ground practitioners to engage with the NLP pipeline in order to analyze alternative data sources. Since much of the content on these alternative channels may involve informal language, local dialects, and context-specific references, allowing practitioners to filter, explore, and improve the output of NLP algorithms is particularly important in this setting. While original research is of vital importance, the use of out-of-the-box tools in NLP pipelines can avoid the duplication of efforts and provide operationally-ready interfaces for practitioners to interact with, which we discuss in further detail below. Some out-of-the box tools have already implemented such functionality for social media sources such as Twitter, as we discuss in further detail below.


\section{Out-of-the-Box Tools for Infodemic Management }\label{sect:tools}
Social listening is an important means for gaining fast insights into public perceptions and information needs as well as detecting misinformation. This is especially important during crises like the COVID-19 pandemic, where there is an overwhelming volume of data, intertwined with conflicting information~\cite{Ruggiero2014Crisis,sheppard2011}. In response to these growing needs, many commercially available machine learning-driven monitoring tools have been developed to public health practitioners, information and communications practitioners and researchers~\cite{Batrinca2014Social}. 
These out-of-the-box tools built by different companies vary in purpose, data input, capabilities, models, workflow management and output~\cite{Stavrakantonakis2012AnAF}. However, while these tools present tremendous value, they also come with novel risks.

Crises strain resources, and sufficient financial, time and personnel resources are not always available to local public health authorities to invest in social media strategies to begin with~\cite{Avery2017Zika}. Devoting staff time to processing, analyzing and real-time monitoring of abundant social media data from scratch can be a luxury to organizations that lack the necessary resources and skills~\cite{Lindsay2011SocialMA}. 
Out-of-the-box tools can 
simplify the data gathering, cleaning, analysis and presentation processes to low- or no-code~\cite{Lee2017social}, providing user-friendly interfaces that facilitate more efficient storage and analysis of data, and offering AI-powered automation for sentiment analysis and continuous monitoring. While in some cases these tools can provide end-to-end insights, they can also be used by technical experts to form one component of a custom-designed NLP pipeline.

Nevertheless, these tools are not without shortcomings. First, the integrity of social media metrics and the ambiguity of their definitions are obstacles for both data analysts and the audience of their reports. `Visible' social media metrics such as the numbers of likes and followers are inherently fallible due to the lack of context (e.g. `liking' a post can be driven by many different motives and emotions) as well as the commercialization of fake followers and metrics ~\cite{Baym2013Data, Cresci2015Fame}. The further processed metrics presented in social media listening tools, such as `reach' and `impressions', suffer from another layer of uncertainty. Often the tools do not fully disclose how the metrics are being calculated, and the definition of each metric may not be standardized across tools, making it even harder for users and readers 
to contextualize the information.

Furthermore, the accuracy of built-in classifiers for sentiment analysis is questionable~\cite{Rappaport2010Listening}. Studies that compare manual sentiment coding with automatic sentiment analysis have shown that there is not only a poor correlation between the manual coding and the tool-generated results~\cite{Deiner2019measles}, but also a lack of agreement between the tools~\cite{Boukes2019tone, young2012affective}. 
It is challenging for researchers to independently investigate why this disagreement occurs, as the proprietary nature of the out-of-the-box tools prevents any external assessment of the integrity of their models. 
This is further complicated by the fact that language models trained prior to COVID-19 might not fully reflect the current reality because of the emergence of Out-of-Vocabulary words~\cite{zheng2020using}.

Some tools attempt to address these problems by implementing human-in-the-loop machine learning, enabling users to correct wrongly classified information~\cite{Stavrakantonakis2012AnAF} or train custom classifiers adapted to users' own context. This highlights the value of versatility in out-of-the-box tools -- predicting changes in the global landscape of communications might not be always possible, but by making these tools more adaptable by design, it is possible to cater to more diverse, novel use cases.

\section{Importance of Partnerships}\label{sect:partnerships}
To unlock the full potential of NLP methodologies and tools to address crises such as COVID-19, it is important that these techniques are appropriately tailored to each particular application setting; 
this should be part of any system design thinking. Partnerships with domain experts, end users and other beneficiaries can serve as a crucial resource to 
help direct research and 
to ensure meaningful and effective deployment. 

In the case of using NLP for the COVID-19 response, productive partnerships could include stakeholders with technical and contextual expertise, data providers and tool developers. In many research settings, methods are developed first and possible applications found later. While this approach is useful for developing theoretical concepts and frameworks, in order to maximize social impact in crisis response scenarios the aims of the application should drive research, with meaningful objectives jointly determined by researchers and implementing partners. Giving greater agency to users of the tools will encourage long-lasting use.

Engaging with partners from different disciplines can be challenging. Often, research agendas and funding cycles require different timelines than the deployment of a tool, sometimes resulting in projects ending prematurely. In the case of many recent projects to tackle COVID-19, it has been difficult to quickly reach the level of maturity needed to operationalize projects and systems at scale~\cite{luengo2020artificial}. It is therefore essential to establish agendas which accommodate the different time constraints of all stakeholders from the beginning of any project. 

In many of the examples mentioned above, partnership development can help NLP researchers, who may be working in geographies which are removed from the application setting of interest, to understand the questions which need to be answered as well as the relevant modalities. For example, in the case of Uganda discussed in Section~\ref{sec:modalities}, local entities could inform researchers that radio streams are a dominant linguistic modality for reliable information and opinion tracking, thereby redirecting effort away from developing social media-related listening strategies. Similarly, partnerships can provide access to relevant data sources, as well as pre-designed tools (see Section \ref{sect:tools}).

\section{Conclusion}
To advance the inclusive use of NLP approaches in the COVID-19 response, we have identified the following focus areas:

\textbf{Data} -- High-quality data helps build high-quality models and guides research priorities and makes it possible to establish benchmarks. 
A major challenge with many of the approaches mentioned above is the need for more inclusive large-scale public datasets to allow NLP researchers to explore and work with low-resource languages and alternative modalities. There are promising endeavors to curate datasets of misinformation such as the Fakeddit dataset~\cite{nakamura2019r}, \href{https://www.poynter.org/ifcn-covid-19-misinformation/}{Poynter} and COVIDLies~\cite{hossain2020covidlies}, as well as as large community modeling efforts such as the \href{https://forums.fast.ai/t/language-model-zoo-gorilla/14623/1}{FastAI Model Zoo}, which could be expanded to cover a wider range of regions and languages. Similarly, efforts to build datasets from a wider range of sources -- e.g., via the creation of radio transcription pipelines -- could encourage researchers to work with data from alternative modalities. Partnerships can also serve as a key source for localized and relevant datasets. Finally, evaluation would be facilitated by the creation of multilingual COVID-related datasets,
as well as a set of benchmarks for how well models perform on certain tasks (e.g. detecting misinformation).

 \textbf{Adaptation and Flexibility} --  There are many existing NLP tools, models and approaches that can be (re-)used in the context of pandemic response. While these tools bring convenience and efficiency to social listening in a crisis like COVID-19, one size does not fit all. Different tools are designed for different purposes and use cases, and modifications are often necessary to adapt them to specific contexts. It is necessary to evaluate the quality of the tools and their algorithms as much as possible to avoid making decisions based on flawed data or analysis~\cite{rai2019explainable}. These tools should be seen as supports for decision making and cannot be relied upon alone; manual validation of the results in context is always crucial~\cite{Boukes2019tone}. For instance, one potential strategy for addressing the limitations of these tools involves the use of human-in-the-loop machine learning to ensure human oversight of ultimate decision-making. 
 
 \textbf{Incentives and Frameworks} -- Beyond the specific context of the pandemic response, the research opportunities described above present challenges that are both academically well-motivated (as many understudied research questions remain to be explored)
 and likely to make a real-world impact (given the urgency and unprecedented nature of the pandemic).
 Even as the market for NLP tools for high-resource languages is becoming increasingly saturated, digital inclusion is creating new markets within communities that currently have few or no customized NLP resources and a large demand for them. There is a need to translate between technical and domain expertise and to bridge the gap between current technological capabilities and the needs of stakeholders.
  Many resources exist to assist in developing data-driven partnerships across different domains~\cite{undp2016guide, dseg2020framework} -- for instance, the UN Sustainable Development Goals and their associated targets can act as a common framework for setting aims and objectives, which is shared among institutions on a global level and can guide the response to and recovery from the COVID-19 pandemic~\cite{unsdgs, vinuesa2020role}. As such, they can assist in the coordination and prioritization of tasks, help with the development of a common language between institutions, and outline effective areas for contributions by the NLP community. 


\newpage
\bibliography{bibliography}
\bibliographystyle{splncs04}

%
%
%
%
%

\end{document}